%% file: main.tex
\documentclass{article} 
\usepackage{iclr2027_conference,times}

\input{math_commands.tex}

\usepackage{hyperref}
\usepackage{url}
\usepackage{amsmath,amssymb}
\usepackage{booktabs}
\usepackage{multirow}
\usepackage{graphicx}
\usepackage{xcolor}
\usepackage{algorithm}
\usepackage{algorithmic}
\usepackage[capitalize,noabbrev]{cleveref}
\usepackage{enumitem}
\usepackage[table]{xcolor}

\newcommand{\chunk}{H}
\newcommand{\exe}{h}
\newcommand{\gate}{\tau}
\newcommand{\method}{Action Upcycling}
\newcommand{\pio}{$\pi_{0.5}$}

\title{Don't Throw Away the Tail: \\ Action Upcycling for Policy Acceleration}

\author{
\textbf{Taesung Kwon}\textsuperscript{\textmd{1*}} \hspace{1em}
\textbf{Jangho Park}\textsuperscript{\textmd{1*}} \hspace{1em}
\textbf{Sunwoo Park}\textsuperscript{\textmd{1}} \hspace{1em}
\textbf{Youngmin Kim}\textsuperscript{\textmd{1}} \\
\ \textbf{Seonghyun Jin}\textsuperscript{\textmd{1}} \hspace{1.3em}
\textbf{Youngjun Jun}\textsuperscript{\textmd{1}} \hspace{0.8em}
\textbf{Kyumin Choi}\textsuperscript{\textmd{1,2}} \hspace{0.6em}
\textbf{Jong Chul Ye}\textsuperscript{\textmd{1}} \\[3pt]
\ \textsuperscript{1}KAIST \hspace{1em} \textsuperscript{2}Sungkyunkwan University
}

\iclrfinalcopy 
\begin{document}

\maketitle

\let\thefootnote\relax\footnotetext{%
  \textsuperscript{*}Equal contribution.\hfill
  Project page and code are available \href{https://acupcycling.github.io/}{\textcolor{magenta}{here}}.
}

\begin{abstract}
Modern robot policies predict a chunk of future actions from a single observation, execute only a prefix, and discard the rest before replanning.
Choosing the length of this prefix, the execution horizon, poses a trade-off between reactivity and efficiency.
A short horizon keeps the policy reactive to the environment, but requires frequent policy calls.
Recent test-time methods adaptively select the horizon for each chunk, but they either read model internals, where the signal must be chosen for each architecture, or draw extra samples, which adds cost.
We propose \emph{Action Upcycling}, a training-free algorithm that reuses actions the policy would otherwise discard, without accessing model internals or drawing extra samples.
We find that discarded actions stay close to their replanned versions as long as the action velocity remains smooth.
Action Upcycling therefore extends the execution horizon up to the point where the velocity begins to fluctuate.
Extensive experiments on simulated and real-world manipulation tasks show that Action Upcycling reduces policy calls by 1.2--1.7$\times$ with no loss in success rate, across multiple Vision-Language-Action Models (VLAs) and even a World Action Model (WAM).
It applies to any chunked policy at negligible cost and is orthogonal to other policy acceleration methods such as few-step sampling and streaming action decoding, opening a new axis for policy acceleration.
\end{abstract}

\section{Introduction}
\label{sec:intro}
Action chunking~\citep{zhao2023learning, chi2025diffusion} has become an essential ingredient in modern robotic imitation learning.
Instead of predicting one action per observation, a policy predicts a sequence of $H$ actions, executes the first $h \le H$ of them in an open-loop manner, and then replans from a fresh observation.
We call $H$ the \emph{prediction horizon} and $h$ the \emph{execution horizon}.
This recipe is shared by modern VLA models such as SmolVLA~\citep{shukor2025smolvla}, GR00T~\citep{bjorck2025gr00t}, and the $\pi$ series~\citep{black2024pi, black2025pi, ai2026pi}.
Recent analyses~\citep{lazzati2026does, zeng2026revisiting} explain why action chunking helps by showing that demonstrated actions are non-Markovian, so a motion is better predicted as a whole than one action at a time.
For example, human demonstrations carry intent across many steps, such as slowing down before contact or pausing at decision boundaries.
Predicting a single action cannot reproduce this behavior, whereas predicting a sequence restores the missing temporal correlation.
Chunking also reduces compounding error and amortizes the latency of a large backbone over many control steps~\citep{black2025real}.

In practice, most policies do not execute the full action chunk but only a prefix of length $h$.
A short $h$ keeps the policy reactive, since it re-observes the environment more frequently and can adjust its behavior in response to execution errors or unexpected changes in the scene.
However, each re-observation requires a full forward pass of a large backbone, so a short $h$ increases the cost.
Furthermore, the optimal value of $h$ varies across tasks and even within a single task~\citep{wang2026vla}.
Only recently have test-time methods~\citep{wang2026vla, xu2026knowing, liang2026adaptive, chen2026dynamic} proposed adaptive selection of $h$ for each action chunk instead of fixing it.
They differ mainly in the signal they read.
\emph{Attention-based} methods track how attention of the action expert is distributed across the chunk~\citep{wang2026vla, xu2026knowing}.
This requires looking inside the model, and which denoising steps, layers, and attention types to read must be chosen for each architecture.
\emph{Sampling-based} methods draw several candidate chunks and measure where they disagree~\citep{liang2026adaptive, chen2026dynamic}.
This requires multiple forward passes, and each additional pass increases the cost.

\begin{figure}[t]
    \centering
    \includegraphics[width=\linewidth]{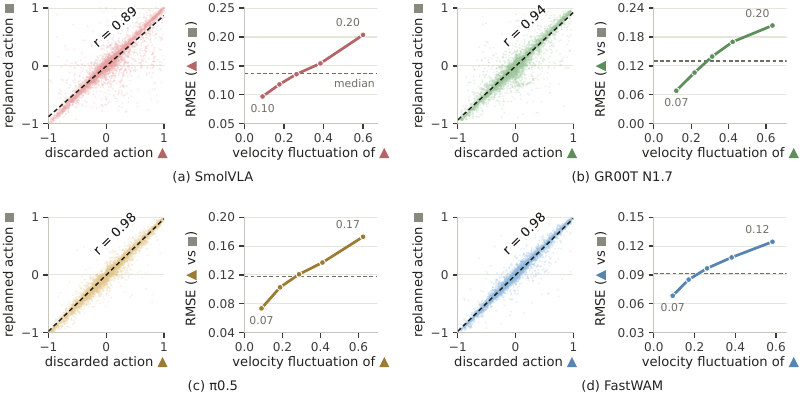}
    \vspace{-0.3cm}
    \caption{\textbf{Discarded actions are worth keeping when motion is smooth.}
    We compare the discarded actions of each chunk with their replanned versions on LIBERO for (a)~SmolVLA, (b)~GR00T N1.7, (c)~$\pi_{0.5}$, and (d)~FastWAM.
    (\emph{left}) They are strongly correlated.
    (\emph{right}) Their RMSE grows with the velocity fluctuation of the discarded actions.
    \emph{Action Upcycling} exploits this property to decide how many discarded actions to reuse.}
    \label{fig:motivation}
\end{figure}

In this work, we propose \emph{Action Upcycling}, a training-free algorithm that reuses actions the policy would otherwise discard, without accessing model internals or drawing extra samples.
It starts from a key observation.
As shown in Figure~\ref{fig:motivation}~(\emph{left}), the discarded actions are strongly correlated with their replanned versions (Pearson correlation between 0.89 and 0.98 across all four policies).
Figure~\ref{fig:motivation}~(\emph{right}) further shows that the discarded actions stay close to them as long as velocity remains smooth.
Action Upcycling is designed to exploit this property, extending the execution horizon up to the point where the predicted velocity begins to fluctuate.
Since the signal is computed from the action chunk that has already been sampled, Action Upcycling is applicable to any chunked policy at negligible cost and is orthogonal to other policy acceleration methods such as few-step sampling and streaming action decoding~\citep{li2026flashvla}, opening a new axis for policy acceleration.

We evaluate Action Upcycling on four policies (SmolVLA, $\pi_{0.5}$, GR00T N1.7, FastWAM) across three benchmarks (LIBERO~\citep{liu2023libero}, LIBERO-Plus~\citep{fei2025libero}, RoboTwin~2.0~\citep{chen2025robotwin}), as well as in real-world robot manipulation tasks.
Across diverse experiments, it consistently reduces the number of policy calls by 1.2--1.7$\times$ with no loss in success rate.
We also demonstrate that Action Upcycling can be seamlessly combined with other policy acceleration methods such as few-step sampling and streaming action decoding.
Notably, it works even on a World Action Model without any modification or tuning, supporting its generality beyond VLAs.

Our contributions are:
\begin{itemize}
    \item We show that discarded actions are strongly correlated with their replanned versions and stay close as long as the action velocity remains smooth.
    \item We propose Action Upcycling, a training-free algorithm that exploits this property to reuse actions the policy would otherwise discard, without accessing model internals or drawing extra samples.
    \item Across four policies (three VLAs and a World Action Model), three benchmarks, and real-world robot manipulation tasks, Action Upcycling consistently reduces policy calls by 1.2--1.7$\times$ with no loss in success rate and combines with other policy acceleration methods.
\end{itemize}

\section{Related Work}
\label{sec:related}
\paragraph{Action chunking and the execution horizon.}
Predicting a chunk of $\chunk$ future actions per policy call was popularized by ACT and Diffusion Policy~\citep{zhao2023learning, chi2025diffusion} and is now standard in VLAs such as SmolVLA, GR00T, and the $\pi$ series~\citep{shukor2025smolvla, bjorck2025gr00t, black2024pi, black2025pi}.
Recent analyses attribute its benefit mainly to the non-Markovian nature of demonstrated actions and reduced compounding error~\citep{lazzati2026does, zeng2026revisiting}.
However, a policy executing a chunk in an open-loop manner does not see the environment until the chunk ends.
BID~\citep{liu2025bidirectional} formalizes this as a trade-off, in which longer execution improves long-term consistency but reduces short-term reactivity.
Therefore, most policies execute only the first $\exe \le \chunk$ actions and discard the rest in practice.
Action Upcycling reuses actions the policy would otherwise discard, reducing unnecessary policy calls and thereby accelerating the policy.

\paragraph{Adaptive execution horizons.}
A concurrent line of work replaces the fixed $\exe$ with an adaptive execution horizon for each action chunk.
These methods differ mainly in the signal used to set the horizon.
\emph{Attention-based} methods read attention weights inside the model.
AutoHorizon~\citep{wang2026vla} uses self-attention of the action expert, and Knowing When to Stop~\citep{xu2026knowing} uses the entropy of cross-attention between action and visual tokens.
These methods require access to model internals, and which layers, attention types, and denoising steps to read must be chosen for each architecture.
\emph{Sampling-based} methods obtain their signal from multiple samples.
AAC~\citep{liang2026adaptive} estimates the entropy across the samples and sets the horizon where the entropy increases most rapidly.
A3~\citep{chen2026dynamic} also samples multiple chunks, selects their medoid as a representative chunk, and executes it only up to the point verified by an additional forward pass.
These methods require additional forward passes, which increase the computational cost.
Unlike these methods, Action Upcycling reads its signal only from a single action chunk, requiring no model internals or extra samples, making it applicable to any chunked policy at negligible cost.

\paragraph{Policy acceleration.}
Most acceleration methods reduce the cost of each policy call, for example by distilling the sampler into fewer denoising steps~\citep{prasad2024consistency, wang2025onestep}, pruning visual tokens or caching intermediate computations~\citep{liu2025bridging, xu2025vla, yang2025efficientvla}, or skipping unnecessary layers of the backbone~\citep{yue2024deer}.
FlashVLA~\citep{li2026flashvla} further accelerates inference by streaming actions from a buffer of chunks at different noise levels.
Action Upcycling reduces the number of policy calls rather than the cost of each call, so the two directions are complementary.
We verify this by combining Action Upcycling with both few-step sampling and FlashVLA in \Cref{sec:experiments}.

\section{Action Upcycling}
\label{sec:method}

\begin{figure}[t]
    \centering
    \includegraphics[width=\linewidth]{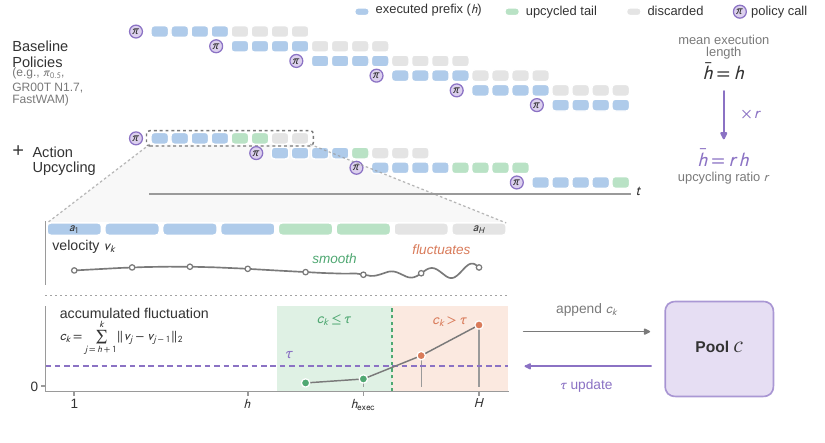}
    \caption{\textbf{Overview of \method{}.}
    \emph{Top}: A chunked policy predicts $H$ actions per call, executes only the first $h$, and discards the rest (the \emph{tail}).
    \method{} additionally executes the trustworthy part of the tail, raising the mean execution length from $h$ to $rh$ and reducing policy calls accordingly.
    \emph{Bottom}: The accumulated velocity fluctuation $c_k$ stays small while the predicted motion is smooth and rises once it fluctuates.
    \method{} executes tail actions while $c_k\le\gate$, where $\gate$ is determined from a pool $\mathcal{C}$ of signals $c_k$, so that the mean execution length reaches $rh$.}
    \label{fig:method}
\end{figure}

\method{} builds on the observation that discarded actions stay close to their replanned versions as long as the predicted velocity remains smooth (\cref{fig:motivation}).
Based on this observation, it decides how many additional actions to execute before replanning, using only the predicted action chunk.
In the following, we first formalize the problem (\cref{sec:method:setup}), and then describe the two components of the method (\cref{fig:method}): a velocity fluctuation signal (\cref{sec:method:signal}) and an adaptive horizon selection strategy that converts the signal into an execution horizon (\cref{sec:method:gate}).

\subsection{Problem Formulation}
\label{sec:method:setup}
In the $i$-th policy call, the policy $\pi$ predicts a chunk of $H$ future actions from the observation $o_{t_i}$ received at control step $t_i$:
\begin{equation}
    A_i \;=\; \pi(o_{t_i}) \;=\; (a_1,\dots,a_H).
    \label{eq:chunk}
\end{equation}
The controller executes the prefix $a_1,\dots,a_h$ and discards the remaining actions $a_{h+1},\dots,a_H$, which we call the \emph{tail}.
\method{} reuses the trustworthy part of the tail, so that the $i$-th call executes $h+\Delta h_i$ actions, where $\Delta h_i\in[0,H-h]$ is the number of reused tail actions.
We define the \emph{mean execution length} $\bar{h}$ as the average number of executed actions per call, and the \emph{upcycling ratio} $r$ as its ratio to $h$:
\begin{equation}
    \bar{h} \;=\; \frac{1}{n}\sum_{i=1}^{n}\left(h+\Delta h_i\right),
    \qquad
    r \;=\; \frac{\bar{h}}{h} \;\in\; [1,\,H/h],
    \label{eq:ratio}
\end{equation}
where $n$ is the total number of policy calls.
For an episode that requires a given number of actions, a larger $r$ proportionally reduces the number of policy calls.
For example, $r=2$ halves the number of policy calls, and $r=1$ recovers the default setting of each policy.
The number of reused tail actions $\Delta h_i$ is determined by where the predicted velocity within the chunk begins to fluctuate, as described in \cref{sec:method:signal,sec:method:gate}.

\subsection{Velocity Fluctuation as a Signal}
\label{sec:method:signal}
Let $v_k$ denote the velocity induced by the $k$-th action in the chunk.
Its definition depends on the action parameterization, with $v_k=a_k$ for relative displacements (e.g.,~LIBERO) and $v_k=a_k-a_{k-1}$ for absolute joint positions (e.g.,~RoboTwin~2.0).
We measure how much the velocity fluctuates beyond the execution horizon $h$ by accumulating the change between consecutive velocities:
\begin{equation}
    c_k \;=\; \sum_{j=h+1}^{k}\big\lVert v_{j}-v_{j-1}\big\rVert_2 ,
    \qquad h< k\le H ,
    \label{eq:cum}
\end{equation}
with $c_h=0$.
Since every summand is non-negative, $c_k$ is non-decreasing along the tail.
It stays small while the predicted motion is steady and rises as soon as the velocity fluctuates.
\cref{fig:motivation} motivates this choice, showing that the RMSE between the discarded actions and their replanned versions grows with the velocity fluctuation.
Therefore, \method{} uses $c_k$ as a signal for how far the tail can be trusted.

\subsection{Adaptive Horizon Selection}
\label{sec:method:gate}
Given a fluctuation threshold $\gate\ge 0$, we reuse tail actions as long as their accumulated fluctuation does not exceed $\gate$, so that the execution length of the chunk is
\begin{equation}
    h_{\text{exec}}(A_i;\gate) \;=\; h+\Delta h_i \;=\; \max_{c_k\le\gate} k ,
    \qquad h\le k\le H ,
    \label{eq:gate}
\end{equation}
where $\Delta h_i$ is the number of reused tail actions.
Setting $\gate=\infty$ executes every chunk as a whole, and $\gate=0$ effectively recovers the default setting of each policy.

Since a larger $\gate$ allows more tail actions to be reused, the upcycling ratio $r$ grows with $\gate$.
This property lets us find the threshold for a given $r$ by a simple search.
Specifically, we collect the velocity fluctuation signals $c_{h+1},\dots,c_H$ of predicted chunks into a pool $\mathcal{C}$, which requires no additional rollouts since $c_k$ depends only on the chunk itself.
We then set $\gate$ to the smallest element of $\mathcal{C}$ for which the mean execution length over the pool satisfies $\bar{h}(\gate)\ge r\,h$.
The pool $\mathcal{C}$ can be built from rollouts of the policy itself, without demonstrations or success labels.
Further analysis of pool construction, including the required pool size and the collection strategy, is provided in \cref{app:analysis:pool}.
The complete \method{} pipeline is summarized in \cref{alg:upcycling}.

\begin{algorithm}[t]
\caption{\method{}}
\label{alg:upcycling}
\begin{algorithmic}[1]
\REQUIRE policy $\pi$, execution horizon $h$, prediction horizon $H$, upcycling ratio $r$
\STATE $\mathcal{C}\leftarrow$ signals $c_{h+1},\dots,c_H$ of chunks predicted by $\pi$
       \hfill $\triangleright$ Pool construction, \cref{sec:method:gate}
\STATE $\gate\leftarrow$ smallest element of $\mathcal{C}$ with $\bar h(\gate)\ge r\,h$
       \hfill $\triangleright$ Threshold update, \cref{sec:method:gate}
\WHILE{task not done}
    \STATE $A=(a_1,\dots,a_H)\leftarrow\pi(o_t)$
    \STATE $c_k\leftarrow\sum_{j=h+1}^{k}\lVert v_{j}-v_{j-1}\rVert_2$, \quad $h<k\le H$
           \hfill $\triangleright$ Velocity fluctuation, \cref{eq:cum}
    \STATE $h_{\text{exec}}\leftarrow\max_{c_k\le\gate}k$, \quad $h\le k\le H$
           \hfill $\triangleright$ Adaptive horizon selection, \cref{eq:gate}
    \STATE execute $a_1,\dots,a_{h_{\text{exec}}}$, \quad $t\leftarrow t+h_{\text{exec}}$
\ENDWHILE
\end{algorithmic}
\end{algorithm}

\section{Experiments}
\label{sec:experiments}
To validate the effectiveness of \method{}, we conduct extensive simulation and real-world experiments.
In the simulation experiments, we evaluate four policies (three VLAs and one World Action Model) on three benchmarks (\cref{sec:exp:sim}).
In the real-world experiments, we deploy two of these policies on a physical robot and evaluate them on diverse manipulation tasks (\cref{sec:exp:real}).

\subsection{Simulation Experiments}
\label{sec:exp:sim}
We apply \method{} to four representative policies, \pio{}~\citep{black2025pi}, SmolVLA~\citep{shukor2025smolvla}, GR00T~N1.7~\citep{bjorck2025gr00t}, and FastWAM~\citep{yuan2026fast}, using their publicly released checkpoints without fine-tuning.
All policies predict a chunk of $\chunk$ actions per call and execute the first $\exe$ actions under their default setting.
The horizons and a detailed description of each policy are provided in \cref{app:details:policies}.
We evaluate on LIBERO~\citep{liu2023libero}, LIBERO-Plus~\citep{fei2025libero}, and RoboTwin~2.0~\citep{chen2025robotwin}, following the official evaluation protocol of each benchmark.
The evaluation covers the four task suites of LIBERO, the seven perturbation dimensions of LIBERO-Plus, and the 50 bimanual manipulation tasks of RoboTwin~2.0 in both clean and randomized settings.
Please refer to \cref{app:details:benchmarks} for detailed descriptions of each benchmark.

\subsubsection{Quantitative Results}
\label{sec:exp:main}
\paragraph{Benefit of \method{}.}
\cref{tab:main} reports the success rate, the number of policy calls per episode, the latency per policy call, and the inference time per episode for four policies on three benchmarks.
Across all combinations, \method{} matches or improves the success rate of the baseline policies while reducing policy calls by 1.2--1.7$\times$ in a training-free manner.
Since the latency per call remains unchanged, the inference time per episode decreases in proportion to the number of calls.
On \pio{}, for instance, \method{} reduces policy calls by 1.4--1.6$\times$ and improves the success rate on all three benchmarks, from 96.9\% to 97.9\% on LIBERO, from 83.7\% to 86.5\% on LIBERO-Plus, and from 59.7\% to 60.5\% on RoboTwin~2.0.
\method{} also works on FastWAM without any modification, reducing its calls by 1.2--1.5$\times$ while improving the success rate, which demonstrates its effectiveness beyond VLAs.

\begin{table}[t]
\vspace{-0.35cm}
\caption{\textbf{Benefit of \method{}.} Results of four policies on three benchmarks. We report success rate (\%), policy calls per episode, latency per policy call (ms), and inference time per episode (s). \method{} consistently reduces policy calls by 1.2--1.7$\times$ while matching or even improving the success rate. Better value in \textbf{bold}.}
\label{tab:main}
\begin{center}
\small
\setlength{\tabcolsep}{4pt}
\resizebox{0.9\linewidth}{!}{%
\begin{tabular}{llcccccccc}
\toprule
 & & \multicolumn{4}{c}{Baseline Policy} & \multicolumn{4}{c}{+ \method{}} \\
\cmidrule(lr){3-6}\cmidrule(lr){7-10}
Benchmark & Model & succ. $\uparrow$ & calls / ep $\downarrow$ & ms & s / ep $\downarrow$ & succ. $\uparrow$ & calls / ep $\downarrow$ & ms & s / ep $\downarrow$ \\
\midrule
\multirow{4}{*}{LIBERO}
 & \pio{} & 96.9 & 32.4 (1$\times$) & 136 & 4.41 & \textbf{97.9} & \textbf{21.9 (1.5$\times$)} & 137 & \textbf{3.00} \\
 & SmolVLA & 82.6 & 19.3 (1$\times$) & 94 & 1.82 & \textbf{82.8} & \textbf{14.7 (1.3$\times$)} & 94 & \textbf{1.39} \\
 & GR00T N1.7 & 96.1 & 22.6 (1$\times$) & 115 & 2.59 & \textbf{96.6} & \textbf{15.0 (1.5$\times$)} & 115 & \textbf{1.72} \\
 & FastWAM & 97.4 & 15.7 (1$\times$) & 84 & 1.32 & \textbf{97.6} & \textbf{13.4 (1.2$\times$)} & 84 & \textbf{1.12} \\
\midrule
\multirow{4}{*}{LIBERO-Plus}
 & \pio{} & 83.7 & 38.5 (1$\times$)& 137 & 5.27 & \textbf{86.5} & \textbf{23.9 (1.6$\times$)} & 137 & \textbf{3.27} \\
 & SmolVLA & 31.8 & 29.4 (1$\times$) & 94 & 2.77 & \textbf{33.3} & \textbf{24.1 (1.2$\times$)} & 94 & \textbf{2.27} \\
 & GR00T N1.7 & 82.1 & 36.4 (1$\times$) & 115 & 4.18 & \textbf{82.3} & \textbf{22.7 (1.6$\times$)} & 115 & \textbf{2.61} \\
 & FastWAM & 49.4 & 33.4 (1$\times$) & 84 & 2.79 & \textbf{49.6} & \textbf{22.5 (1.5$\times$)} & 84 & \textbf{1.88} \\
\midrule
\multirow{3}{*}{RoboTwin 2.0}
 & \pio{} & 59.7 & 40.9 (1$\times$) & 152 & 6.22 & \textbf{60.5} & \textbf{29.8 (1.4$\times$)} & 152 & \textbf{4.53} \\
 & SmolVLA & 34.8 & 50.6 (1$\times$) & 114 & 5.77 & \textbf{47.4} & \textbf{29.7 (1.7$\times$)} & 114 & \textbf{3.39} \\
 & FastWAM & 89.5 & 10.6 (1$\times$) & 214 & 2.27 & \textbf{90.6} & \textbf{8.7 (1.2$\times$)} & 214 & \textbf{1.86} \\
\bottomrule
\end{tabular}}
\end{center}
\end{table}

\paragraph{Comparison with adaptive execution horizon methods.}
We compare \method{} with two representative adaptive execution horizon methods on \pio{}, the attention-based AutoHorizon~\citep{wang2026vla} and the sampling-based AAC~\citep{liang2026adaptive}.
\cref{tab:unified} reports the success rate, the number of policy calls per episode, the latency per policy call, and the inference time per episode on three benchmarks.
On LIBERO and LIBERO-Plus, \method{} achieves the highest success rate, the largest call reduction, and the shortest inference time per episode, without access to model internals.
AAC reduces policy calls, but it draws multiple samples per call, which increases both the latency per call and the inference time per episode.
On RoboTwin~2.0, AutoHorizon reduces policy calls slightly more but degrades the success rate, and AAC even increases policy calls, whereas \method{} improves the success rate while still reducing policy calls.

\begin{table}[t]
\vspace{-0.3cm}
\caption{\textbf{Comparison with adaptive execution horizon methods.} Results of \pio{} on three benchmarks. We report success rate (\%), policy calls per episode, latency per policy call (ms), and inference time per episode (s). Best in \textbf{bold}, second best \underline{underlined}.}
\vspace{-0.2cm}
\label{tab:unified}
\begin{center}
\small
\setlength{\tabcolsep}{4pt}
\resizebox{\linewidth}{!}{%
\begin{tabular}{l cccc cccc cccc}
\toprule
 & \multicolumn{4}{c}{LIBERO} & \multicolumn{4}{c}{LIBERO-Plus} & \multicolumn{4}{c}{RoboTwin~2.0} \\
\cmidrule(lr){2-5}\cmidrule(lr){6-9}\cmidrule(lr){10-13}
Method & succ. & calls / ep & ms & s / ep & succ. & calls / ep & ms & s / ep & succ. & calls / ep & ms & s / ep \\
\midrule
Baseline & 96.9 & 32.4 (1$\times$) & 136 & 4.41 & 83.7 & 38.5 (1$\times$) & 137 & 5.27 & \underline{59.7} & 40.9 (1$\times$) & 152 & 6.22 \\
+ AAC (CVPR 2026) & 97.3 & 23.5 (1.38$\times$) & 802 & 18.9 & 84.5 & 27.1 (1.42$\times$) & 802 & 21.7 & 53.1 & 46.1 (0.89$\times$) & 930 & 42.9 \\
+ AutoHorizon (ECCV 2026) & \underline{97.4} & \underline{22.3 (1.45$\times$)} & 136 & \underline{3.03} & \underline{85.5} & \underline{24.5 (1.57$\times$)} & 136 & \underline{3.33} & 56.8 & \textbf{28.0 (1.46$\times$)} & 152 & \textbf{4.26} \\
\rowcolor{gray!25}
+ \method{} & \textbf{97.9} & \textbf{21.9 (1.48$\times$)} & 137 & \textbf{3.00} & \textbf{86.5} & \textbf{23.9 (1.61$\times$)} & 137 & \textbf{3.27} & \textbf{60.5} & \underline{29.8 (1.37$\times$)} & 152 & \underline{4.53} \\
\bottomrule
\end{tabular}}
\end{center}
\end{table}

\paragraph{Combination with other policy acceleration methods.}
Most policy acceleration methods reduce the cost of each policy call, whereas \method{} reduces the number of calls, so the two directions are complementary.
We combine \method{} with two such methods (few-step sampling and streaming action decoding with FlashVLA~\citep{li2026flashvla}) on \pio{}.
\cref{tab:stack} shows that, even with fewer denoising steps ($N{=}5$ and $N{=}2$), \method{} keeps the call reduction at about 1.5$\times$ and improves the success rate over the baseline with the same $N$.
As a result, combining \method{} with few-step sampling raises the total speed-up to 2.8$\times$ at $N{=}2$.
\cref{tab:stack_flashvla} shows that FlashVLA reduces the latency per policy call from 136~ms to 32~ms.
Building on this, \method{} reduces the number of policy calls by 1.51$\times$ while improving the success rate.
As a result, combining \method{} with FlashVLA yields a total speed-up of 6.7$\times$.
Implementation details of all compared and combined methods are provided in \cref{app:details:baselines}.

\begin{table}[t]
\vspace{-0.35cm}
\caption{\textbf{\method{} stacks with denoising-step reduction.} Results of \pio{} with $N$ denoising steps on LIBERO. We report success rate (\%), policy calls per episode, latency per policy call (ms), inference time per episode (s), and speed-up over the $N{=}10$ baseline. Better value at each $N$ in \textbf{bold}.}
\label{tab:stack}
\begin{center}
\small
\setlength{\tabcolsep}{4pt}
\resizebox{0.6\linewidth}{!}{%
\begin{tabular}{l ccccc}
\toprule
Setting & succ. $\uparrow$ & calls / ep $\downarrow$ & ms & s / ep $\downarrow$ & speed-up $\uparrow$ \\
\midrule
Baseline ($N{=}10$) & 96.9 & 32.4 (1$\times$) & 136 & 4.41 & 1.0$\times$ \\
\rowcolor{gray!25}
+ \method{} & \textbf{97.9} & \textbf{21.9 (1.48$\times$)} & 137 & \textbf{3.00} & \textbf{1.5$\times$} \\
\midrule
$N{=}5$ & 96.9 & 32.4 (1$\times$) & 97 & 3.14 & 1.4$\times$ \\
\rowcolor{gray!25}
+ \method{} & \textbf{97.7} & \textbf{21.9 (1.48$\times$)} & 98 & \textbf{2.15} & \textbf{2.1$\times$} \\
\midrule
$N{=}2$ & 96.8 & 32.0 (1$\times$) & 69 & 2.21 & 2.0$\times$ \\
\rowcolor{gray!25}
+ \method{} & \textbf{97.8} & \textbf{21.8 (1.47$\times$)} & 71 & \textbf{1.55} & \textbf{2.8$\times$} \\
\bottomrule
\end{tabular}}
\end{center}
\end{table}

\begin{table}[t]
\vspace{-0.5cm}
\caption{\textbf{\method{} stacks with streaming action decoding.} Results of \pio{} with FlashVLA on LIBERO. FlashVLA denoises a rolling buffer of chunks at different noise levels for faster decoding. We report success rate (\%), policy calls per episode, latency per policy call (ms), inference time per episode (s), and speed-up over \pio{}. Best in \textbf{bold}.}
\label{tab:stack_flashvla}
\begin{center}
\small
\setlength{\tabcolsep}{4pt}
\resizebox{0.75\linewidth}{!}{%
\begin{tabular}{l c c c c c}
\toprule
Setting & succ. $\uparrow$ & calls / ep $\downarrow$ & ms & s / ep $\downarrow$ & speed-up $\uparrow$ \\
\midrule
Baseline & 96.90 & 32.4 (1$\times$) & 136 & 4.41 & 1.0$\times$ \\
+ FlashVLA  & 98.65 & 31.4 (1.03$\times$) & 32 & 1.01 & 4.4$\times$ \\
\rowcolor{gray!25}
+ FlashVLA \& \method{} & \textbf{98.70} & \textbf{21.5 (1.51$\times$)} & 31 & \textbf{0.66} & \textbf{6.7$\times$} \\
\bottomrule
\end{tabular}}
\vspace{-0.2cm}
\end{center}
\end{table}

\subsubsection{Ablation Study}
\label{sec:exp:ablation}

\begin{figure}[h!]
\centering
\vspace{-0.2cm}
\includegraphics[width=\linewidth]{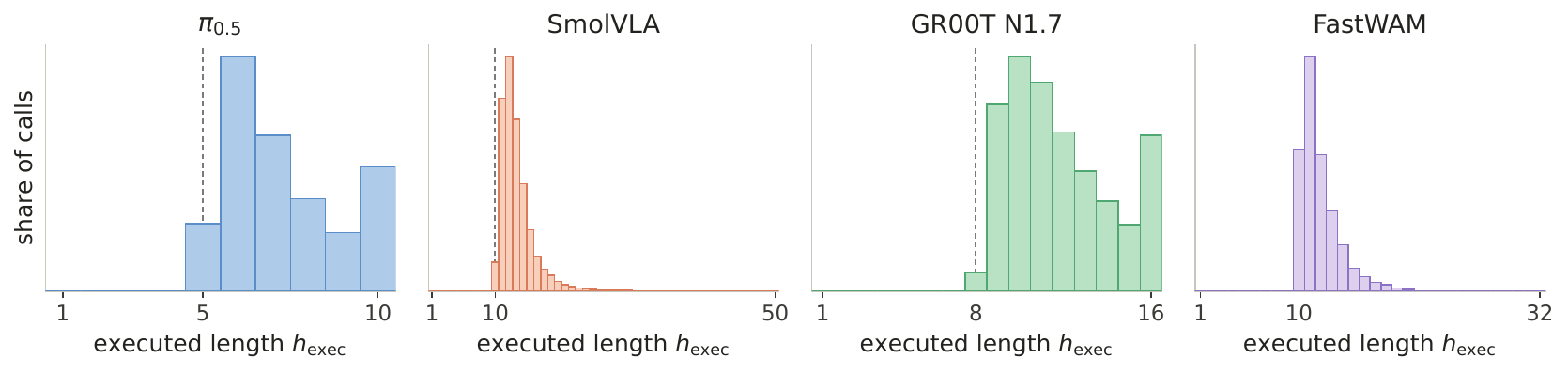}
\vspace{-0.7cm}
\caption{\textbf{Distribution of execution lengths.} The execution lengths selected by \method{} spread over a range rather than concentrating on a single value.}
\label{fig:behaviour}
\end{figure}

One may attribute the gain of \method{} simply to executing more actions per call.
To examine this possibility, we replace our adaptive execution with fixed-length execution, where the length is set close to the mean execution length $\bar{\exe}$ of \method{}.
\cref{tab:controls_libero} shows that fixed-length execution degrades the success rate by 0.7--1.1 points compared to the baseline for three out of four policies.
In contrast, with a similar number of policy calls, our adaptive execution consistently improves the success rate by 0.9--1.3 points over fixed-length execution across all four policies.
Furthermore, \cref{fig:behaviour} shows that the execution lengths selected by \method{} do not concentrate on a single value but spread over a range.
This indicates that \method{} dynamically adapts the number of tail actions to execute for each chunk based on the velocity fluctuation signal.
Further analysis of how each policy behaves under different upcycling ratios is provided in \cref{app:analysis:ratio}.

\begin{table}[h!]
\vspace{-0.1cm}
\caption{\textbf{Adaptive vs.\ fixed-length execution.} Results of four policies on LIBERO. Fixed-length execution uses a length close to the mean execution length $\bar{\exe}$ of \method{}. We report success rate (\%) and $\bar{\exe}$. Better value in \textbf{bold}.}
\label{tab:controls_libero}
\begin{center}
\small
\setlength{\tabcolsep}{4pt}
\resizebox{0.55\linewidth}{!}{%
\begin{tabular}{lcccccc}
\toprule
 & \multicolumn{2}{c}{Baseline} & \multicolumn{2}{c}{Fixed-length} & \multicolumn{2}{c}{Adaptive (ours)} \\
\cmidrule(lr){2-3}\cmidrule(lr){4-5}\cmidrule(lr){6-7}
Model & succ. $\uparrow$ & $\bar{\exe}$ & succ. $\uparrow$ & $\bar{\exe}$ & succ. $\uparrow$ & $\bar{\exe}$ \\
\midrule
\pio{}              & 96.90 & 5.0  & 97.05 & 7.0  & \textbf{97.90} & 7.3  \\
SmolVLA             & 82.60 & 10.0 & 81.55 & 13.0 & \textbf{82.80} & 13.3 \\
GR00T N1.7          & 96.05 & 8.0  & 95.40 & 12.0 & \textbf{96.55} & 11.9 \\
FastWAM             & 97.35 & 10.0 & 96.45 & 12.0 & \textbf{97.55} & 11.9 \\
\bottomrule
\end{tabular}}
\end{center}
\vspace{-0.3cm}
\end{table}

\subsection{Real-World Experiments}
\label{sec:exp:real}
We evaluate \method{} on real-world robot manipulation tasks to verify that its effectiveness carries over to physical execution.
We use \pio{} and GR00T~N1.6 as base policies and deploy them on the YAM arm, a 6-DoF single-arm manipulator with a linear gripper.
We describe the data collection, policy training, and evaluation protocol below.

\paragraph{Data collection.}
We collect 424 teleoperated episodes on the YAM arm with a leader arm, covering eight tasks with about 53 episodes each.
The tasks consist of four pick-and-place tasks and four long-horizon drawer tasks that require opening a drawer, moving an object, and closing the drawer.
Each episode records a top-down camera view, a wrist camera view, the joint state, and the action, represented as absolute joint positions, at 30~Hz.

\paragraph{Policy training.}
We fine-tune both policies on the collected data with their official training recipes, using 20{,}000 steps with a batch size of 64 for \pio{} and 10{,}000 steps with a batch size of 32 for GR00T~N1.6, while keeping all other settings at their defaults.
Both policies are trained to predict absolute joint positions as actions, with a prediction horizon of $\chunk${=}50 actions for \pio{} and $\chunk${=}16 actions for GR00T~N1.6.

\paragraph{Evaluation.}
We evaluate \pio{} on all eight tasks and GR00T~N1.6 on the four pick-and-place tasks, with 10 episodes per task.
The default execution horizon is $\exe${=}15 for \pio{} and $\exe${=}9 for GR00T~N1.6, which gave the most stable execution in our preliminary trials.
Since actions are absolute joint positions, we compute $v_k$ as the difference of consecutive joint positions, excluding the gripper.
We set $\gate$ using $r${=}1.25 for \pio{} and $r${=}1.1 for GR00T~N1.6.
The baseline and \method{} share the same checkpoint and control settings.
An episode counts as a success only if every step of the instruction is completed, with no partial credit.
Further details on the robot setup and computing resources are provided in \cref{app:details:real,app:details:computing}.

\subsubsection{Quantitative Results}
\cref{tab:realworld,tab:realworld_groot} report success rate, policy calls per episode, and time per episode for each task with \pio{} and GR00T~N1.6.
For both policies, \method{} matches or improves the success rate on every task while reducing policy calls.

For \pio{}, the overall success rate increases from 72/80 to 77/80, and policy calls decrease from 48.3 to 34.0 per episode.
On the pick-and-place tasks, where the baseline already succeeds in every episode, \method{} maintains the perfect success rate while reducing policy calls by 1.30$\times$ on average.
On the long-horizon drawer tasks, \method{} even improves the success rate from 32/40 to 37/40 and reduces policy calls by 1.48$\times$.

For GR00T~N1.6, the overall success rate slightly increases from 34/40 to 35/40, and policy calls decrease from 90.6 to 71.0 per episode.
The time per episode also decreases for both policies, from 24.9~s to 21.5~s for \pio{} and from 28.0~s to 24.9~s for GR00T~N1.6, indicating that the policy call reduction from \method{} translates into faster task completion in real-world deployment.

\begin{table}[h]
\caption{\textbf{Quantitative real-world results with \pio{}.} Each task is evaluated over 10 episodes. We report success rate, policy calls per episode, and time per episode (s). Better value in \textbf{bold}.}
\label{tab:realworld}
\begin{center}
\small
\setlength{\tabcolsep}{4pt}
\resizebox{\linewidth}{!}{%
\begin{tabular}{lcccccc}
\toprule
 & \multicolumn{2}{c}{Success rate $\uparrow$} & \multicolumn{2}{c}{Calls / ep $\downarrow$} & \multicolumn{2}{c}{Time / ep (s) $\downarrow$} \\
\cmidrule(lr){2-3}\cmidrule(lr){4-5}\cmidrule(lr){6-7}
Task & Baseline & \method{} & Baseline & \method{} & Baseline & \method{} \\
\midrule
Soccer ball $\rightarrow$ plate (layout 1) & 10/10 & 10/10 & 26.9 & \textbf{26.8} & \textbf{14.4} & 17.4 \\
Soccer ball $\rightarrow$ plate (layout 2) & 10/10 & 10/10 & 36.5 & \textbf{20.7} & 19.0 & \textbf{12.9} \\
Basketball $\rightarrow$ basket            & 10/10 & 10/10 & 31.3 & \textbf{24.9} & 16.5 & \textbf{15.8} \\
Baseball $\rightarrow$ basket              & 10/10 & 10/10 & 32.1 & \textbf{25.0} & 17.1 & \textbf{16.1} \\
Drawer: black ball $\rightarrow$ drawer    &  8/10 & \textbf{10/10} & 73.8 & \textbf{49.2} & 37.3 & \textbf{30.9} \\
Drawer: basketball $\rightarrow$ drawer    &  8/10 &  8/10 & 61.2 & \textbf{40.7} & 31.6 & \textbf{25.5} \\
Drawer: baseball $\rightarrow$ basket      &  7/10 & \textbf{9/10} & 67.9 & \textbf{40.6} & 34.3 & \textbf{25.5} \\
Drawer: basketball $\rightarrow$ pot       &  9/10 & \textbf{10/10} & 56.9 & \textbf{44.5} & 28.7 & \textbf{27.8} \\
\midrule
Total / mean                               & 72/80 & \textbf{77/80} & 48.3 & \textbf{34.0} & 24.9 & \textbf{21.5} \\
\bottomrule
\end{tabular}}
\end{center}
\end{table}

\begin{table}[h]
\vspace{-0.35cm}
\caption{\textbf{Quantitative real-world results with GR00T~N1.6.} Each task is evaluated over 10 episodes. We report success rate, policy calls per episode, and time per episode (s). Better value in \textbf{bold}.}
\label{tab:realworld_groot}
\begin{center}
\small
\setlength{\tabcolsep}{4pt}
\resizebox{\linewidth}{!}{%
\begin{tabular}{lcccccc}
\toprule
 & \multicolumn{2}{c}{Success rate $\uparrow$} & \multicolumn{2}{c}{Calls / ep $\downarrow$} & \multicolumn{2}{c}{Time / ep (s) $\downarrow$} \\
\cmidrule(lr){2-3}\cmidrule(lr){4-5}\cmidrule(lr){6-7}
Task & Baseline & \method{} & Baseline & \method{} & Baseline & \method{} \\
\midrule
Soccer ball $\rightarrow$ plate (layout 1) & 10/10 & 10/10 & 56.8 & \textbf{50.3} & 17.8 & \textbf{17.7} \\
Soccer ball $\rightarrow$ plate (layout 2) &  8/10 & \textbf{9/10} & 102.2 & \textbf{81.7} & 31.5 & \textbf{28.8} \\
Basketball $\rightarrow$ basket            &  8/10 &  8/10 & 95.6 & \textbf{69.9} & 29.6 & \textbf{24.3} \\
Baseball $\rightarrow$ basket              &  8/10 &  8/10 & 107.9 & \textbf{82.3} & 33.1 & \textbf{28.9} \\
\midrule
Total / mean                               & 34/40 & \textbf{35/40} & 90.6 & \textbf{71.0} & 28.0 & \textbf{24.9} \\
\bottomrule
\end{tabular}}
\vspace{-0.1cm}
\end{center}
\end{table}

\begin{figure}[t]
    \centering
    \includegraphics[width=\linewidth]{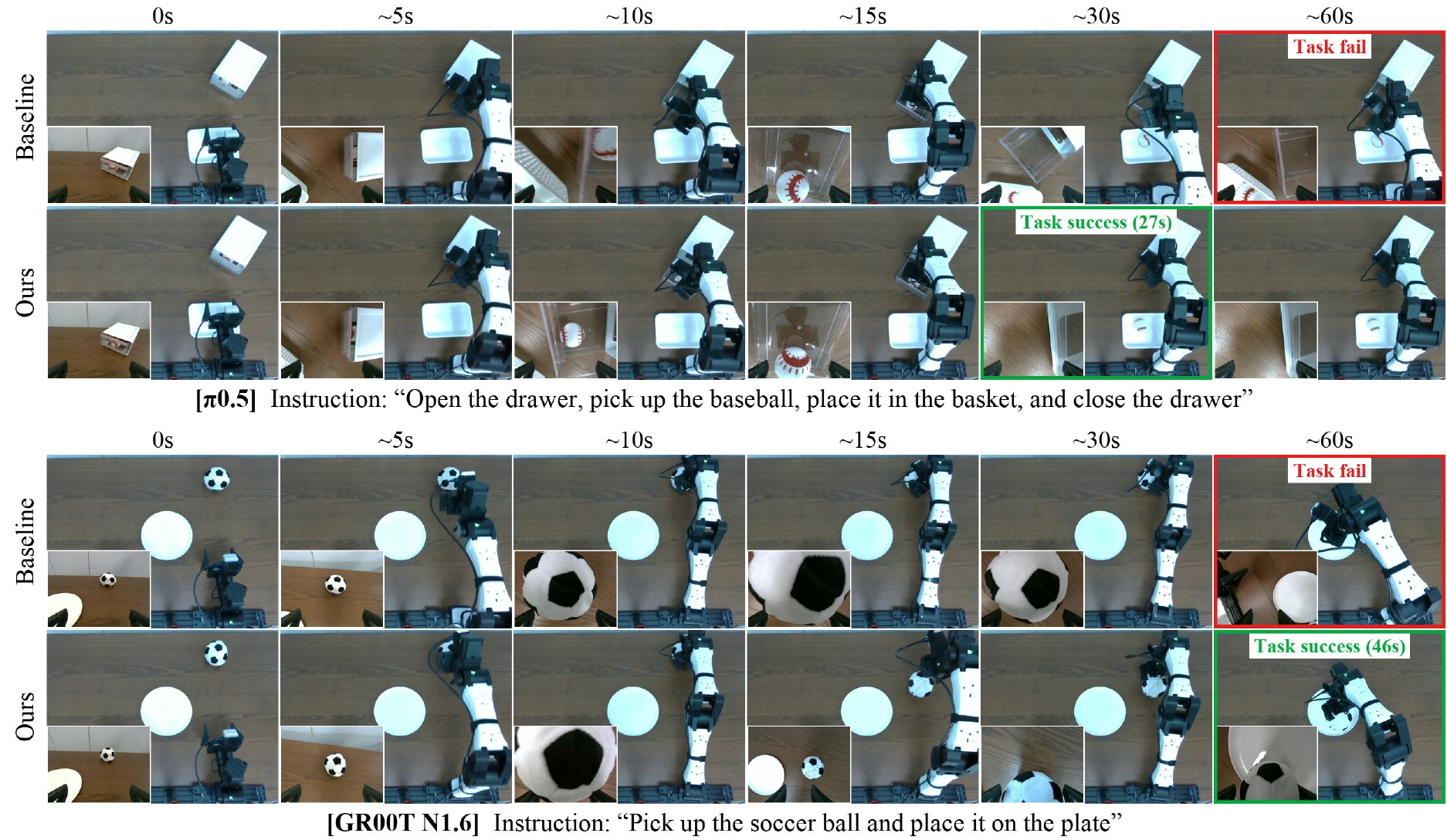}
    \vspace{-0.7cm}
    \caption{\textbf{Qualitative real-world results.} A drawer task with \pio{} (top) and a pick-and-place task with GR00T~N1.6 (bottom), each shown with the baseline policy and with \method{}. Red and green boxes mark failure and success, respectively. In both tasks, the baseline policy times out before completion, while the same policy with \method{} succeeds. Videos are available on the project page.}
    \vspace{-0.1cm}
    \label{fig:realworld}
\end{figure}

\subsubsection{Qualitative Results}
\cref{fig:realworld} shows real-world rollouts of \pio{} on a drawer task and of GR00T~N1.6 on a pick-and-place task.
In both tasks, the baseline policy times out at 60~s before completion, whereas the same policy with \method{} succeeds in 27~s and 46~s, respectively.
In episodes where the baseline policy also succeeds, the same policy with \method{} tends to complete the task faster, as shown in \cref{app:real_qualitative}.
Videos of the real-world rollouts are available on the project page.

\section{Conclusion}
\label{sec:conclusion}
We proposed \method{}, a simple but effective algorithm that reuses the tail of each action chunk, which chunked policies usually discard.
We showed that discarded actions stay close to replanned versions as long as their motion remains smooth.
We used the accumulated velocity fluctuation as a signal for how far the tail can be trusted.
Since the signal is computed from a single predicted chunk, \method{} requires no access to model internals, no extra samples, and no additional training.
Across three VLAs and a World Action Model, three benchmarks, and real-world experiments, \method{} demonstrates its effectiveness and robustness in policy acceleration and composes with orthogonal acceleration methods, opening a new axis for faster policy inference.

\clearpage

\section*{Reproducibility statement}
We describe the proposed method in \cref{sec:method} and provide further implementation details in \cref{app:additional_details}.
Since our method is training-free, all simulation results can be reproduced with publicly available policy checkpoints and benchmarks.
We also release the implementation of our method.
Although the real-robot data and fine-tuned weights will not be released, we describe the data collection setup and fine-tuning procedure in \cref{app:additional_details}.

\bibliography{iclr2027_conference}
\bibliographystyle{iclr2027_conference}

\clearpage
\appendix

\section{Further Implementation Details}
\label{app:additional_details}

\subsection{Policies}
\label{app:details:policies}
We apply \method{} to the following policies using their released checkpoints without any fine-tuning for simulation experiments.
The execution and prediction horizons of each policy are listed in \cref{tab:app:hh}.

\begin{itemize}
    \item \textbf{\pio{}}~\citep{black2025pi} is a VLA with a flow-matching action expert that attends to vision-language tokens at every layer through shared attention.
    We use $N${=}10 denoising steps.
    For RoboTwin~2.0, we use the public checkpoint released by Motus~\citep{bi2026motus}.
    \item \textbf{SmolVLA}~\citep{shukor2025smolvla} is a compact VLA with a flow-matching action expert that consists of alternating cross-attention and self-attention blocks and attends to a frozen vision-language model.
    We use $N${=}10 denoising steps.
    \item \textbf{GR00T N1.7}~\citep{bjorck2025gr00t} is a VLA with a flow-matching action expert, which receives features from a frozen vision-language model through cross-attention.
    We use $N${=}4 denoising steps.
    \item \textbf{FastWAM}~\citep{yuan2026fast} is a World Action Model that generates actions jointly with future video latents.
    We use $N${=}10 denoising steps.
\end{itemize}

\begin{table}[h]
\caption{\textbf{Horizon settings.} Execution horizon $\exe$ and prediction horizon $\chunk$ of each model and benchmark.}
\label{tab:app:hh}
\begin{center}
\small
\begin{tabular}{lccc}
\toprule
Model & LIBERO & LIBERO-Plus & RoboTwin 2.0 \\
\midrule
\pio{}      & 5 / 10  & 5 / 10  & 10 / 32 \\
SmolVLA     & 10 / 50 & 10 / 50 & 10 / 50 \\
GR00T N1.7  & 8 / 16  & 8 / 16  & --- \\
FastWAM     & 10 / 32 & 10 / 32 & 24 / 32 \\
\bottomrule
\end{tabular}
\end{center}
\end{table}

\subsection{Benchmarks}
\label{app:details:benchmarks}
We follow the official evaluation protocol of each benchmark unless otherwise specified.

\begin{itemize}
    \item \textbf{LIBERO}~\citep{liu2023libero} is a simulated single-arm manipulation benchmark consisting of four suites that focus on spatial layouts, objects, task goals, and long-horizon tasks, with ten tasks each.
    The action is a relative end-effector displacement.
    \item \textbf{LIBERO-Plus}~\citep{fei2025libero} extends LIBERO to evaluate robustness under seven perturbation dimensions, namely camera viewpoint, robot initial state, language, lighting, background, sensor noise, and object layout.
    It shares the action space of LIBERO.
    \item \textbf{RoboTwin~2.0}~\citep{chen2025robotwin} is a bimanual manipulation benchmark consisting of 50 tasks, which we evaluate in both clean and randomized settings.
    The action is an absolute joint position.
\end{itemize}

\subsection{Baselines}
\label{app:details:baselines}
We compare or combine \method{} with the following methods, using their official implementations and default hyperparameters unless otherwise specified.

\begin{itemize}
    \item \textbf{AAC}~\citep{liang2026adaptive} samples multiple chunks per policy call and executes actions up to the point where the entropy across the samples increases most rapidly. We use $K{=}20$ samples and the default hyperparameters.
    \item \textbf{AutoHorizon}~\citep{wang2026vla} determines the execution horizon from the attention of the action expert. We use the official soft pointer implementation.
    \item \textbf{FlashVLA}~\citep{li2026flashvla} advances a rolling buffer of four action chunks by one denoising step per call, thereby reducing the latency of each call.
    We use the \pio{} checkpoint fine-tuned by the authors and execute the same number of actions per call as in the baseline.
\end{itemize}

\subsection{Real-World Experimental Setup}
\label{app:details:real}

\begin{figure}[h]
\centering
\includegraphics[width=0.6\linewidth]{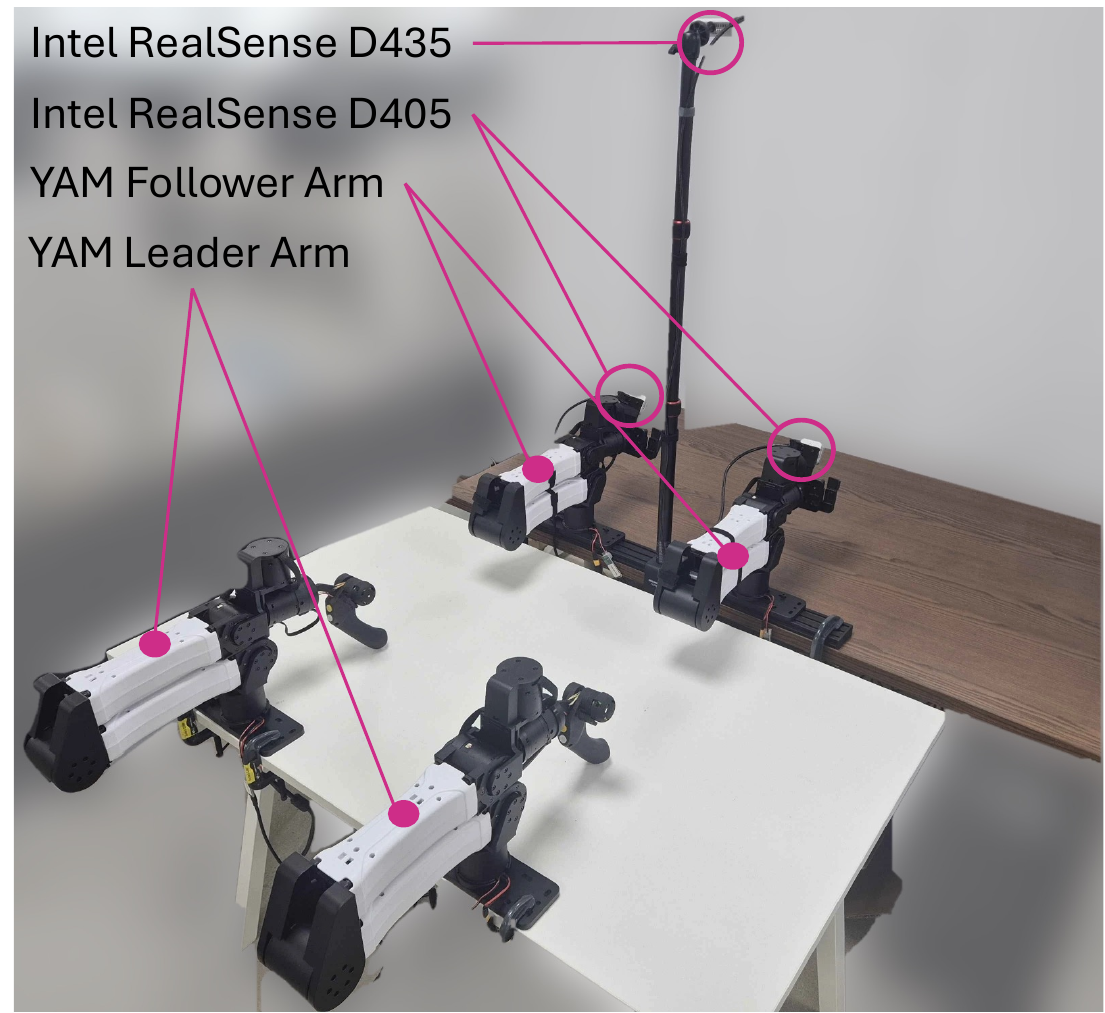}
\caption{\textbf{Real-robot setup.} Two YAM follower arms with wrist-mounted Intel RealSense D405 cameras, a top-mounted Intel RealSense D435, and two YAM leader arms used for teleoperated data collection. All experiments use only the right follower arm.}
\label{fig:real_setup}
\vspace{-0.3cm}
\end{figure}

\paragraph{Robot setup.}
\cref{fig:real_setup} shows the robot station used in our real-world experiments.
It consists of two YAM follower arms, driven over a CAN bus with joint-position PD control, and a matching pair of YAM leader arms for teleoperation.
All experiments in this paper use only the right follower arm.
An Intel RealSense D435 mounted on a pole above the table provides the top-down view, and an Intel RealSense D405 on the wrist of each arm provides the wrist view.
The policy receives the RGB streams from the two cameras at a resolution of 640$\times$480, along with the joint state, consisting of six joint positions and the gripper opening, and outputs targets for the same seven dimensions.
The joint velocity is limited to 0.75~rad/s.
All episodes are stored in the LeRobot format.

\paragraph{Fine-tuning \pio{}.}
We fine-tune the released \pio{} base checkpoint with the official post-training recipe of openpi.
The YAM state and action are 7-dimensional, whereas \pio{} operates on a 32-dimensional state and action space.
Following the recipe, both are zero-padded to 32 dimensions and normalized with per-dimension statistics computed on our dataset, and the padded dimensions are ignored at deployment.
All parameters are updated during fine-tuning. We train for 20{,}000 steps with a batch size of 64, keeping the other settings of the recipe unchanged.

\paragraph{Fine-tuning GR00T N1.6.}
We fine-tune GR00T N1.6 with the official recipe of Isaac-GR00T.
The YAM is registered as a new embodiment, with a modality configuration that declares the state and action fields (six arm joints and the gripper) and the two camera views.
Following the default recipe for new embodiments, the vision-language backbone is kept frozen, and only the embodiment-specific projectors and the diffusion transformer are fine-tuned.
Arm actions are predicted relative to the current joint state, while the gripper action is predicted as an absolute value.
State and action are normalized with per-dimension statistics computed on our dataset.
We train for 10{,}000 steps with a batch size of 32, keeping the other settings of the recipe unchanged.

\subsection{Computing Resources}
\label{app:details:computing}
All simulation experiments are run on NVIDIA RTX 3090 GPUs, with one policy per GPU.
The latency per policy call and the inference time per episode reported in \cref{tab:main,tab:unified,tab:stack,tab:stack_flashvla} are measured on a single NVIDIA RTX 4090 GPU.
For the real-world experiments in \cref{sec:exp:real}, \pio{} and GR00T~N1.6 are fine-tuned on a single NVIDIA B200 GPU and deployed on a single NVIDIA RTX 4090 GPU.

\clearpage

\section{Further Analysis}
\label{app:additional_analysis}

\subsection{Pool Construction}
\label{app:analysis:pool}

The pool $\mathcal{C}$ of velocity fluctuation signals can be built from rollouts of the policy itself, without demonstrations or success labels (\cref{sec:method:gate}).
\Cref{tab:calib} shows that a small pool suffices for selecting $\gate$.
Using the signals of only a random 2\,\% of the chunks yields nearly the same $\gate$ and resulting mean execution length $\bar{h}$ as using all of them, across all four policies.

Furthermore, the pool need not be collected in advance.
While the results above use an offline pool collected before deployment, we can also build an online pool of signals collected during deployment and update $\gate$ accordingly.
As shown in \cref{tab:online_vs_offline}, the online pool achieves a comparable success rate and call reduction to the offline pool.
In practice, we use the offline pool in all experiments, as it provides a slightly higher success rate and a shorter episode time.

\begin{table}[h]
\caption{\textbf{A small pool suffices for selecting $\gate$.} Selected $\gate$ and resulting mean execution length $\bar{h}$ on LIBERO, using pools built from all chunks and from a random 2\,\% of them.}
\label{tab:calib}
\begin{center}
\small
\setlength{\tabcolsep}{6pt}
\begin{tabular}{lcccc}
\toprule
 & \multicolumn{2}{c}{All chunks} & \multicolumn{2}{c}{2\,\% of chunks} \\
\cmidrule(lr){2-3}\cmidrule(lr){4-5}
Model & $\gate$ & $\bar{h}$ & $\gate$ & $\bar{h}$ \\
\midrule
\pio{}      & 0.15 & 7.3  & 0.15 & 7.3  \\
SmolVLA     & 0.65 & 13.3 & 0.64 & 13.2 \\
GR00T N1.7  & 0.22 & 11.9 & 0.22 & 11.9 \\
FastWAM     & 0.09 & 11.9 & 0.09 & 11.9 \\
\bottomrule
\end{tabular}
\end{center}
\end{table}

\begin{table}[h]
\caption{\textbf{Offline vs.\ online pool construction.} Results of \pio{} on LIBERO. Best in \textbf{bold}.}
\label{tab:online_vs_offline}
\begin{center}
\small
\setlength{\tabcolsep}{6pt}
\begin{tabular}{l cccc}
\toprule
Method & succ. $\uparrow$ & calls / ep $\downarrow$ & ms / call $\downarrow$ & s / ep $\downarrow$ \\
\midrule
Baseline & 96.9 & 32.4 (1$\times$) & \textbf{136} & 4.41 \\
+ \method{} (offline pool $\mathcal{C}$) & \textbf{97.9} & 21.9 (1.48$\times$) & 137 & \textbf{3.00} \\
+ \method{} (online pool $\mathcal{C}$) & 97.7 & \textbf{21.0 (1.54$\times$)} & 150 & 3.15 \\
\bottomrule
\end{tabular}
\end{center}
\end{table}

\subsection{Effect of the Upcycling Ratio}
\label{app:analysis:ratio}

\begin{figure}[h]
\centering
\includegraphics[width=\linewidth]{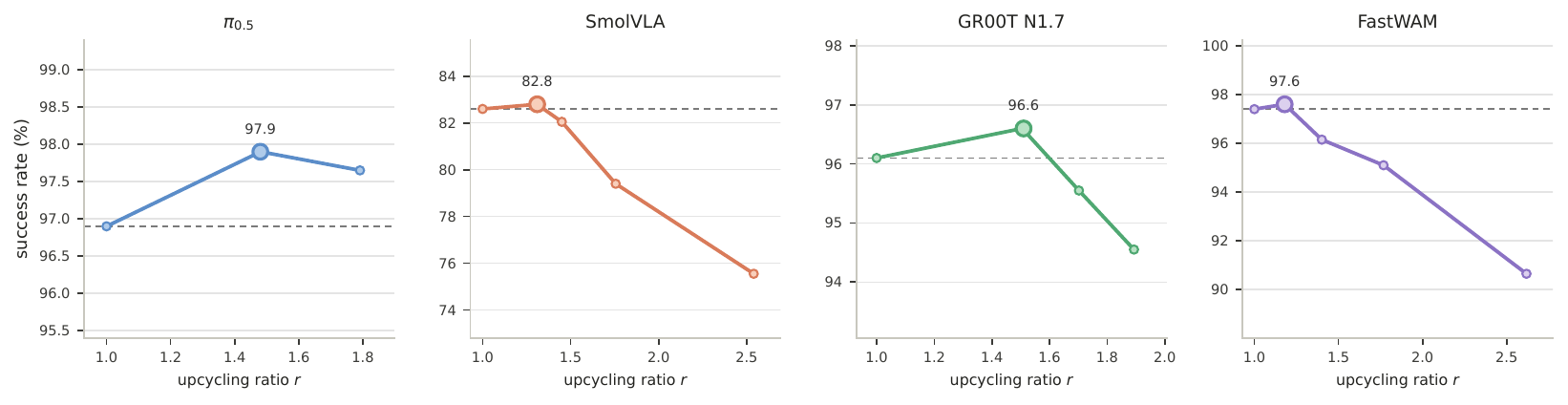}
\caption{\textbf{Effect of the upcycling ratio.} Success rate (\%) of each policy on LIBERO as $r$ increases from the default setting ($r{=}1$).}
\label{fig:sweep}
\end{figure}

\cref{fig:sweep} shows how the success rate of each policy on LIBERO behaves as the upcycling ratio $r$ increases from the default setting ($r{=}1$).
For every policy, there is an upcycling ratio between 1.2 and 1.7 that slightly improves the success rate while reducing policy calls, in line with the main results in \cref{tab:main}.
Beyond this ratio, the success rate gradually drops, likely because a larger $r$ admits tail actions with larger velocity fluctuation.

\clearpage

\section{More Qualitative Real-World Results}
\label{app:real_qualitative}

\begin{figure}[h]
\centering
\includegraphics[width=\linewidth]{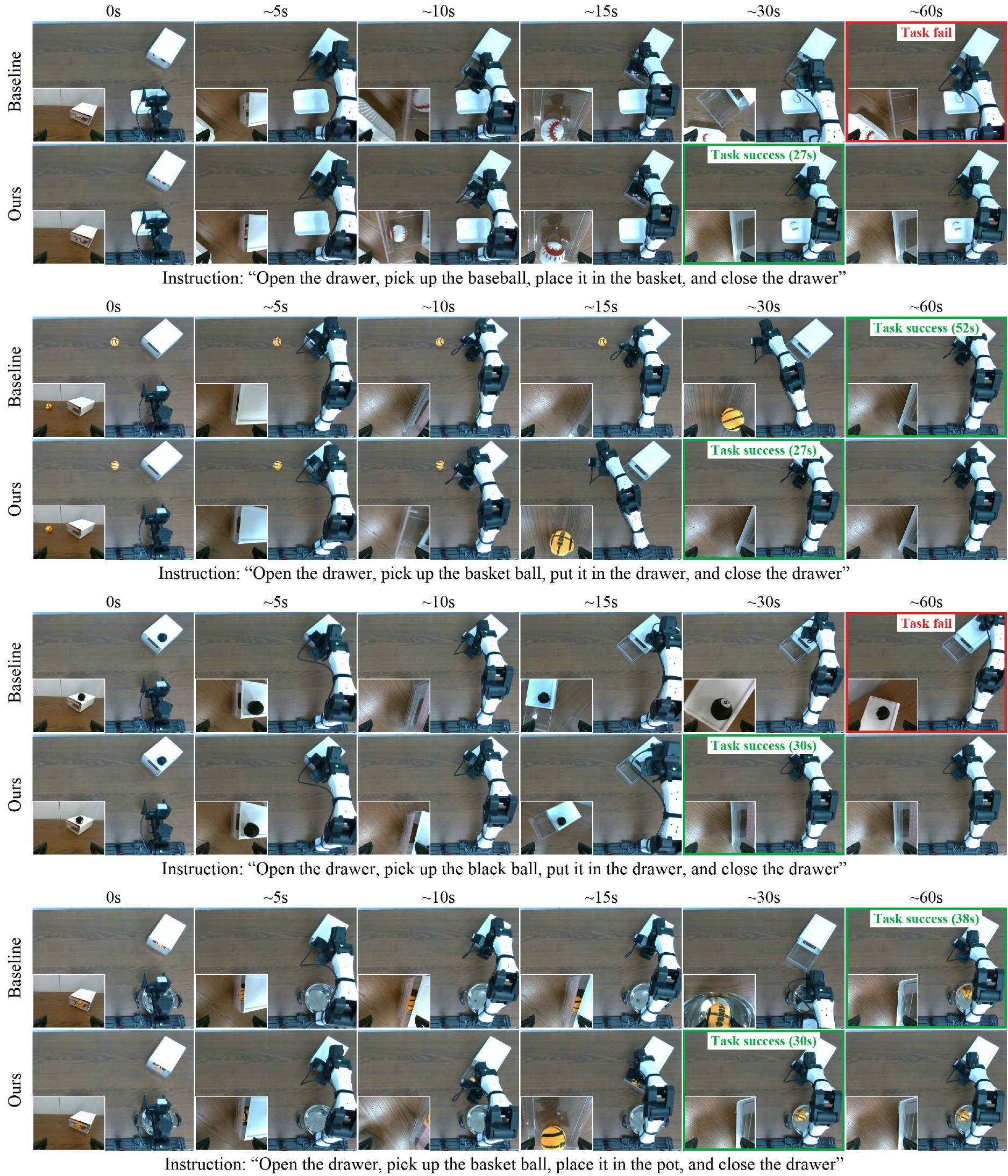}
\caption{\textbf{Additional qualitative real-world results with \pio{}.} Four drawer tasks, each shown with the baseline policy (top) and with \method{} (bottom). Red and green boxes mark failure and success, respectively. In two tasks, the baseline policy times out before completion, while the same policy with \method{} succeeds. In the other two tasks, both succeed, but the policy with \method{} completes the task faster.}
\label{fig:realworld_pi}
\end{figure}

\begin{figure}[h]
\centering
\includegraphics[width=\linewidth]{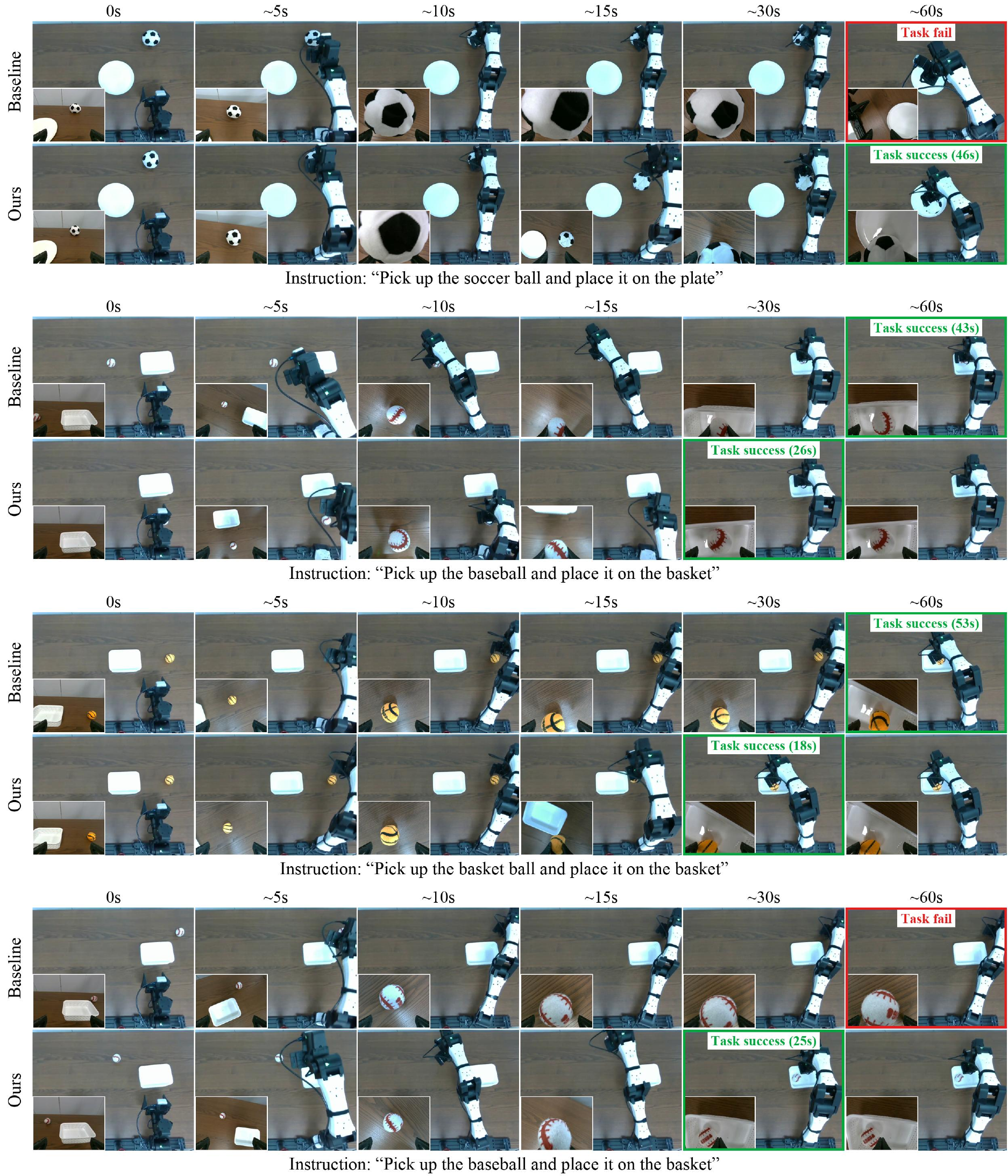}
\caption{\textbf{Additional qualitative real-world results with GR00T~N1.6.} Four pick-and-place tasks, each shown with the baseline policy (top) and with \method{} (bottom). Red and green boxes mark failure and success, respectively. In two tasks, the baseline policy times out before completion, while the same policy with \method{} succeeds. In the other two tasks, both succeed, but the policy with \method{} completes the task faster.}
\label{fig:realworld_groot}
\end{figure}

\end{document}

%% file: math_commands.tex
\usepackage{amsmath,amsfonts,bm}

\def\eqref#1{equation~\ref{#1}}

\def\1{\bm{1}}

\DeclareMathAlphabet{\mathsfit}{\encodingdefault}{\sfdefault}{m}{sl}
\SetMathAlphabet{\mathsfit}{bold}{\encodingdefault}{\sfdefault}{bx}{n}

